\documentclass[runningheads]{llncs}

\usepackage{eccv}

\usepackage{multirow}
\usepackage{algorithm}
\usepackage{algpseudocode}
\usepackage{listings}
\usepackage{tcolorbox}
\usepackage{xcolor}
\usepackage{wrapfig}

\usepackage{eccvabbrv}

\usepackage{graphicx}
\usepackage{booktabs}

\usepackage[accsupp]{axessibility}  

\usepackage{hyperref}

\usepackage{orcidlink}
\usepackage{pdfpages}
\usepackage{marvosym}

\begin{document}

\lstdefinestyle{prompt}{
    basicstyle=\ttfamily\fontsize{7pt}{8pt}\selectfont,
    frame=none,
    breaklines=true,
    backgroundcolor=\color{lightgray},
    breakatwhitespace=true,
    breakindent=0pt,
    escapeinside={(*@}{@*)},
    numbers=none,
    numbersep=5pt,
    xleftmargin=5pt,
    literate={`}{\textasciigrave}1
}
\tcbset{
  aibox/.style={
    top=10pt,
    colback=white,
    colframe=black,
    colbacktitle=black,
    center,
  }
}
\newtcolorbox{AIbox}[2][]{aibox, title=#2,#1}

\newcommand{\redtext}[1]{\textcolor{blue}{#1}}

\title{OmniMapBench: Benchmarking Visual-Centric Reasoning on Diverse Map Documents} 

\titlerunning{OmniMapBench}

\author{Yang Chen\inst{1,2}$^\ast$ \and
Yunwen Li\inst{3}$^\ast$ \and
Yufan Shen\inst{1}$^\ast$ \and
Minghao Liu\inst{3,4}$^\ast$ \and
Tianyu Zheng\inst{3} \and \\ 
Bin Fu\inst{1} \and 
Qunshu Lin\inst{4} \and 
Zhi Yu\inst{2} \and 
Botian Shi\inst{1,5}\textsuperscript{\Letter}}

\authorrunning{Y.~Chen et al.}

\institute{
\textsuperscript{1}Shanghai Artificial Intelligence Laboratory \quad
\textsuperscript{2}Zhejiang University \quad \\
\textsuperscript{3}M-A-P \quad
\textsuperscript{4}2077AI \quad
\textsuperscript{5}Shanghai Innovation Institute \\
\email{zjucheny@gmai.com}
}

\maketitle
\begin{NoHyper}
\def\thefootnote{}\footnotetext{$^\ast$Equal contribution. \textsuperscript{\Letter}Corresponding author.}
\end{NoHyper}

\begin{abstract}

Recent advancements in LVLMs necessitate robust benchmarks for complex, visually grounded reasoning. A critical limitation is identified in many document understanding benchmarks: visual content is often reducible to text, enabling high performance without genuine visual grounding.
To address this limitation, OmniMapBench is introduced to foster visual-centric reasoning for map documents. The benchmark comprises 2,096 manually annotated question-answer pairs across 1,603 map documents from nine categories. It is designed to probe a hierarchy of skills, ranging from perception to multi-step visual reasoning.
To quantify benchmark properties, a simple yet effective benchmark-level metric is proposed: the Visual Dependency Index (VDI), defined as the accuracy drop when images are replaced with question-agnostic descriptions. OmniMapBench exhibits higher VDI than established benchmarks, which quantitatively validates its focus on irreducible visual reasoning. 
Comprehensive evaluations of 25 leading LVLMs are conducted on OmniMapBench. A significant performance gap is observed, with the top-performing model achieving only 75.03\% accuracy. This result underscores the challenges posed by OmniMapBench to current LVLMs.
This work aims to catalyze progress in visual-centric reasoning for document understanding of LVLMs. The dataset and code are publicly available at \url{https://github.com/SIGMME/OmniMapBench}.


\end{abstract}

\section{Introduction}
\label{sec:intro}

\begin{figure}[t]
\centering
\includegraphics[width=0.99\linewidth]{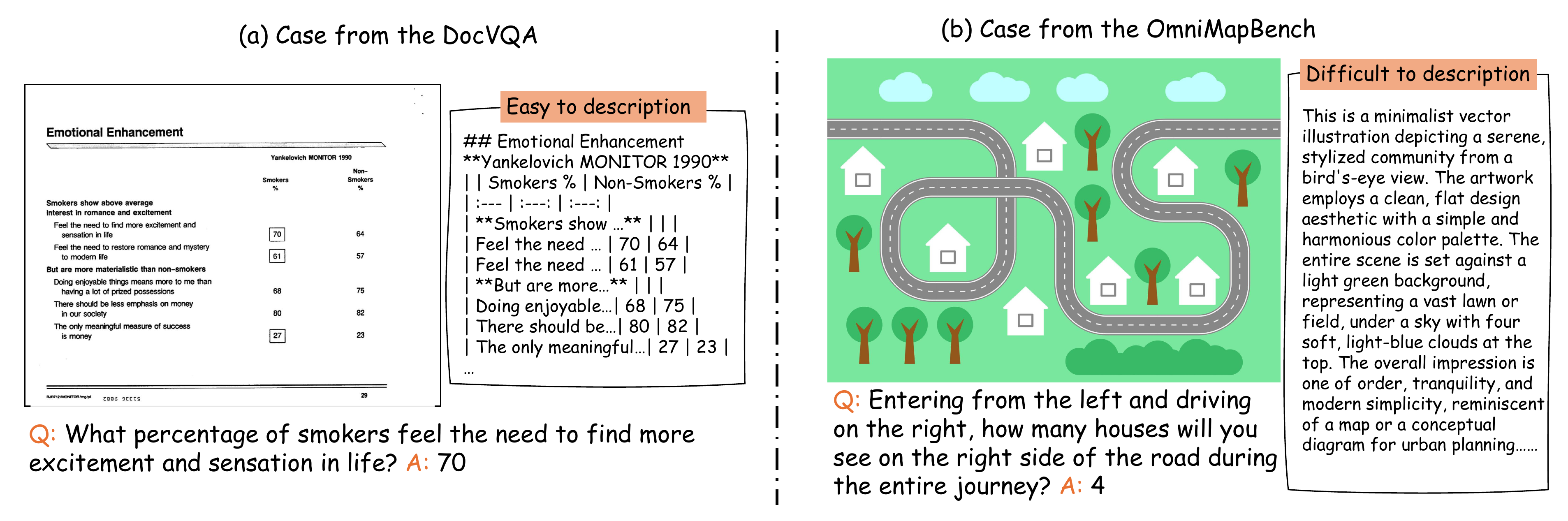}
\caption{Comparison of textualizability across benchmarks. (a) A DocVQA case with structured tabular content is easily described in text, and the question is solvable from the description. (b) In contrast, the \textit{simple} map image from OmniMapBench cannot be sufficiently captured by a question-agnostic textual description to solve the question. The task requires visual-centric reasoning that is irreducible to a linear text format.}
\label{fig:intro}
\end{figure}

\begin{figure}[]
\centering
\includegraphics[width=0.99\linewidth]{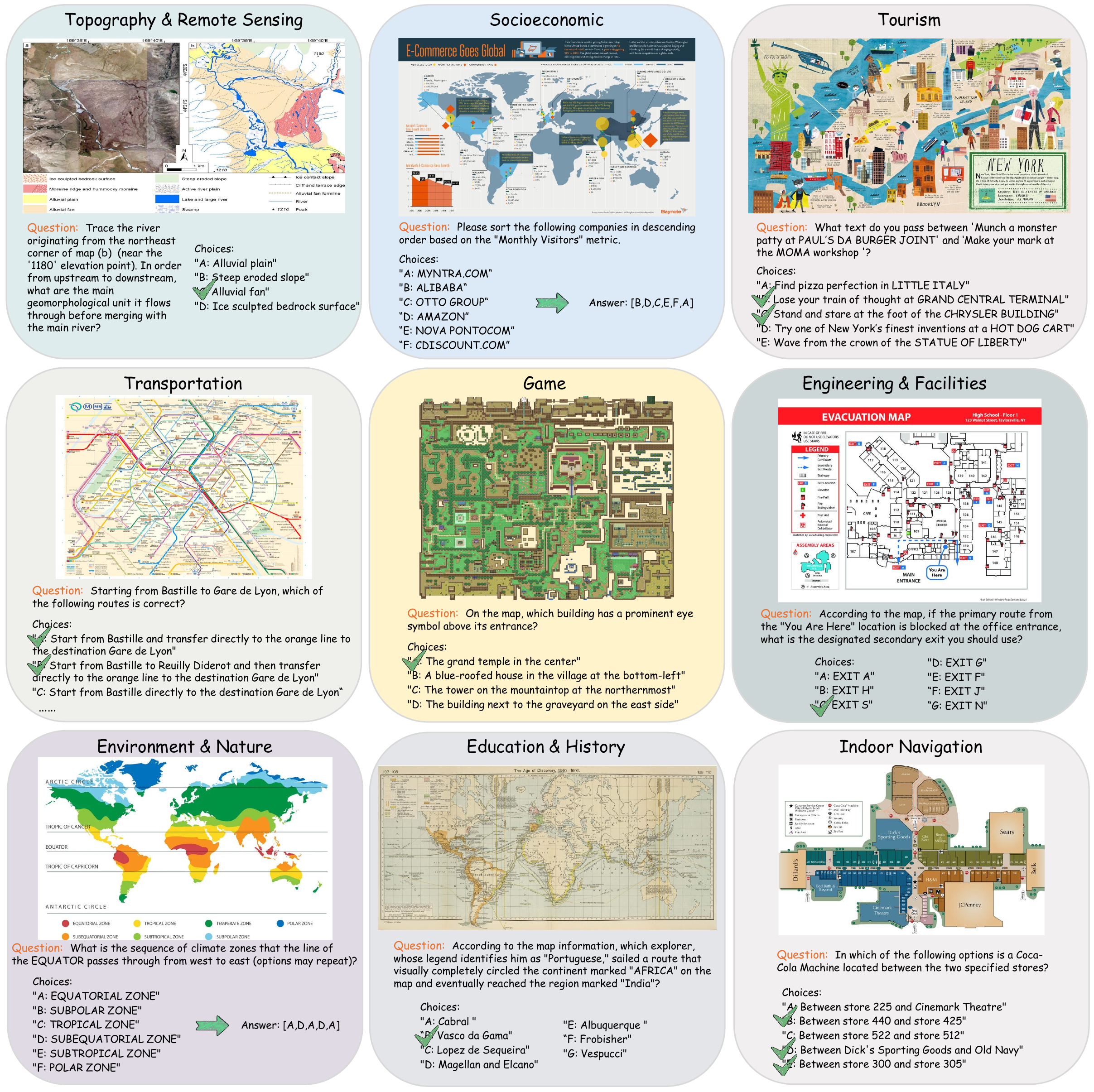}
\caption{Sample questions and answers from OmniMapBench across 9 map categories. The benchmark encompasses diverse visual styles and reasoning tasks, ranging from foundational topographic analysis to fictional virtual worlds  and remote sensing imagery. Questions probe perception, single-step spatial reasoning, and multi-step relational reasoning grounded in map-specific visual cues.}
\label{fig:overview}
\end{figure}

Advances in the reasoning capabilities of Large Language Models (LLMs)~\cite{openaio1,deepseekr1,team2025kimi} have catalyzed the development of Large Vision-Language Models (LVLMs)\cite{qwen3vl,gpt5card,gemini25pro,seed16,claudesonnet4-5}. Consequently, visual reasoning has emerged as a central research area~\cite{wang2025mathcoder,zhao2025chartcoder,suris2023vipergpt,zhang2025chain,huang2025high}. The objective is shifting from perception tasks, such as object recognition and captioning, towards complex inference grounded in visual content. To this end, various ``think with images''\cite{openaio3,su2025openthinkimg,zheng2025deepeyes,su2025thinking} methods are explored, wherein models are trained to decompose intricate visual queries and execute programmatic actions via external tools\cite{chen2025learning,wu2025vtool}. This paradigm facilitates a more adaptive and compositional approach to problem-solving, enabling models to address challenges that transcend simple visual perception.

The rapid evolution of model capabilities, however, places new demands on evaluation benchmarks. As shown in Figure \ref{fig:intro}, a significant limitation is identified in many existing document Visual Question Answering (VQA) benchmarks~\cite{mathew2021docvqa}: the visual content in these benchmarks is often reducible to structured text. For instance, document images can be converted into markdown representations through OCR and layout analysis~\cite{blecher2023nougat,kim2021donut,stanislawek2021kleister,liu2024ocrbench,fu2024ocrbench,chen2026babyvisionvisualreasoninglanguage}, which is a common practice in retrieval-augmented generation (RAG) pipelines. Charts and tables are frequently expressible as code or data structures~\cite{masry2022chartqa,liu2023deplot,mathew2022infographicvqa}. In mathematical and geometric reasoning tasks, the visual diagrams are often text-complete~\cite{liu2024mmbench,yue2024mmmu,yue2025mmmu}, meaning that a verbal description of the geometric configuration suffices for problem solving~\cite{lu2023mathvista,luo2025geogrambench,hiippala2021ai2d}. This reducibility suggests that many existing benchmarks do not require genuine visual grounding and can be solved through text-based inference alone.

To address the need for more visual-centric evaluation, maps are put forth as a document category that inherently resists textual reduction~\cite{feng2026reasonmap}. Unlike documents amenable to OCR~\cite{mathew2021docvqa, lewis2020retrieval} or table~\cite{shen2025proctag} and charts~\cite{masry2022chartqa} deconstructible into codes, the core semantic content of a map is embedded within its spatial topology, symbolic conventions, and continuous graphical features~\cite{li2025mapqa,feng2026reasonmap,xing2025can}. Information is conveyed through the interplay of contour lines, color-coded regions, relational positions, and complex legends that cannot be adequately captured by linear text descriptions. For example, the seemingly simple map in Figure~\ref{fig:intro}(b) illustrates this difficulty: without foreknowledge of the specific query, it is challenging to formulate a textual description that exhaustively encodes all spatial relationships and symbolic details necessary for arbitrary reasoning tasks. Consequently, while human interpretation of a map may feel intuitive, the requisite synthesis of fine-grained perception and multi-step spatial reasoning presents a formidable challenge for current models, making maps an ideal medium for assessing genuine visual-centric reasoning.

This work introduces OmniMapBench, a benchmark designed specifically to target and evaluate visual-centric reasoning. As shown in Figure~\ref{fig:overview}, the benchmark encompasses a diverse collection of maps, organized into 9 categories. This collection ranges from topographic and economic maps to historical and fantasy maps, ensuring high visual diversity. A set of approximately 2,096 question-answer pairs is constructed through a careful manual annotation process. To facilitate straightforward and objective evaluation, the tasks are structured into three formats: single-choice, multiple-choice, and ordering. These tasks are designed to probe a hierarchy of skills, including perception, single-step reasoning, and multi-step reasoning.

To quantitatively assess the reliance of a benchmark on visual information, this work proposes a simple yet effective metric. The Visual Dependency Index (VDI) measures the performance degradation observed when the full visual input is replaced by a text-only question-agnostic description generated by a vision-language model. A high VDI score signifies that a task is difficult to solve through textual shortcuts and thus requires genuine visual reasoning. 

A comprehensive evaluation is conducted on 25 open-source and closed-source LVLMs. The results show that even the state-of-the-art model achieves 75.03\% accuracy on OmniMapBench, indicating a substantial performance gap. Furthermore, experiments using the VDI protocol demonstrate that OmniMapBench possesses a significantly higher VDI compared to several established document VQA benchmarks. This finding quantitatively validates its reduced susceptibility to textualization shortcuts and confirms its focus on visual-centric reasoning.

The main contributions are summarized as follows:

\begin{itemize}

    \item The introduction of OmniMapBench, a diverse, visual-centric benchmark for map understanding featuring 2,096 manually annotated and verified question-answer pairs of 1,603 map images across 9 map categories.
    
    \item The proposal of the Visual Dependency Index (VDI), a simple yet effective metric to quantify a LVLM benchmark's reliance on visual information.

    \item A comprehensive evaluation of 25 LVLMs is conducted, providing a comparison and identifying performance bottlenecks across map categories and visual-centric reasoning capabilities.

\end{itemize}

\section{Related Work}

\subsection{Large Visual Language Models}
Large Visual Language Models (LVLMs) have developed rapidly and have shown significant progress in various visual language tasks. Both open-source models ~\cite{qwen3vl, wang2025internvl35advancingopensourcemultimodal} and powerful proprietary models ~\cite{gpt4vsystemcard,gemini25pro} have pushed the frontiers of artificial intelligence~\cite{seed16,claudesonnet4,coreteam2025mimovltechnicalreport,kimiteam2025kimivltechnicalreport}. Early progress was marked by a substantial improvement in foundational skills, such as optical character recognition (OCR)~\cite{singh2019towards,mathew2021docvqa,jaume2019funsd}, enabling models to read and interpret documents and charts with increasing accuracy~\cite{kim2022ocr,peng2023kosmos,blecher2023nougat,luo2024layoutllm,xu2020layoutlm,xu2020layoutlmv2,huang2022layoutlmv3}. Subsequently, LVLMs mastered compositional abilities like complex spatial instruction following and multi-turn visual dialogue~\cite{liu2023visual,alayrac2022flamingo,das2017visual}, shifting the research frontier towards higher-order visual reasoning~\cite{zellers2019recognition,saikh2022scienceqa}. To address visual-centric reasoning, the research focus is centered on enabling models to ``think with images''~\cite{thinkwithimages,su2025thinking}. Three strategies are studied. In visual synthesis, auxiliary images are generated to support the reasoning process~\cite{chern2025thinking,zhang2025scaling,wu2025reinforcing}. In focused perception, salient regions are selected and zoomed for fine-grained inspection~\cite{openaio3,shen2024zoomeye,zhang2025chain,huang2025high,shao2024visual,zheng2025deepeyes}. In code-based reasoning, executable programs are produced and executed to derive answers~\cite{chen2025learning,suris2023vipergpt,wang2025mathcoder,zhao2025chartcoder}.

\subsection{LVLM Benchmarks}
Evaluations for Large Vision-Language Models ~\cite{yin2024survey,li2024survey} range from foundational understanding~\cite{chen2015microsoft,yu2016modeling} to complex visual reasoning~\cite{park2020visualcomet}. Initial benchmarks established baselines for foundational understanding, such as visual question answering~\cite{antol2015vqa,goyal2017making,marino2019ok} and reading text within images~\cite{singh2019towards,mathew2021docvqa,masry2022chartqa}. Subsequently, the evaluative focus broadened to compositional reasoning, where models are required to interpret relationships within structured data or maintain context across conversational turns~\cite{das2017visual,de2017guesswhat,yang2024chartmimic}. This progression naturally culminated in benchmarks targeting higher-order cognitive reasoning, which probe a model's inferential logic, often by requiring an explicit rationale for an answer~\cite{shao2024visual,wang2503multimodal}, or challenge it with complex, multi-step problems from knowledge-intensive domains like science and mathematics~\cite{lu2023mathvista,yue2024mmmu,zhou2025mdk12}. Current evaluation efforts focus on benchmarks of greater complexity, designed to probe deeper visual-centric reasoning~\cite{jimenez2023swe,chen2025iwr,guo2025r,chen2026babyvisionvisualreasoninglanguage} and test the capability boundaries of SOTA models.

\section{OmniMapBench Dataset}

\begin{figure}[t]
\centering
\includegraphics[width=0.99\linewidth]{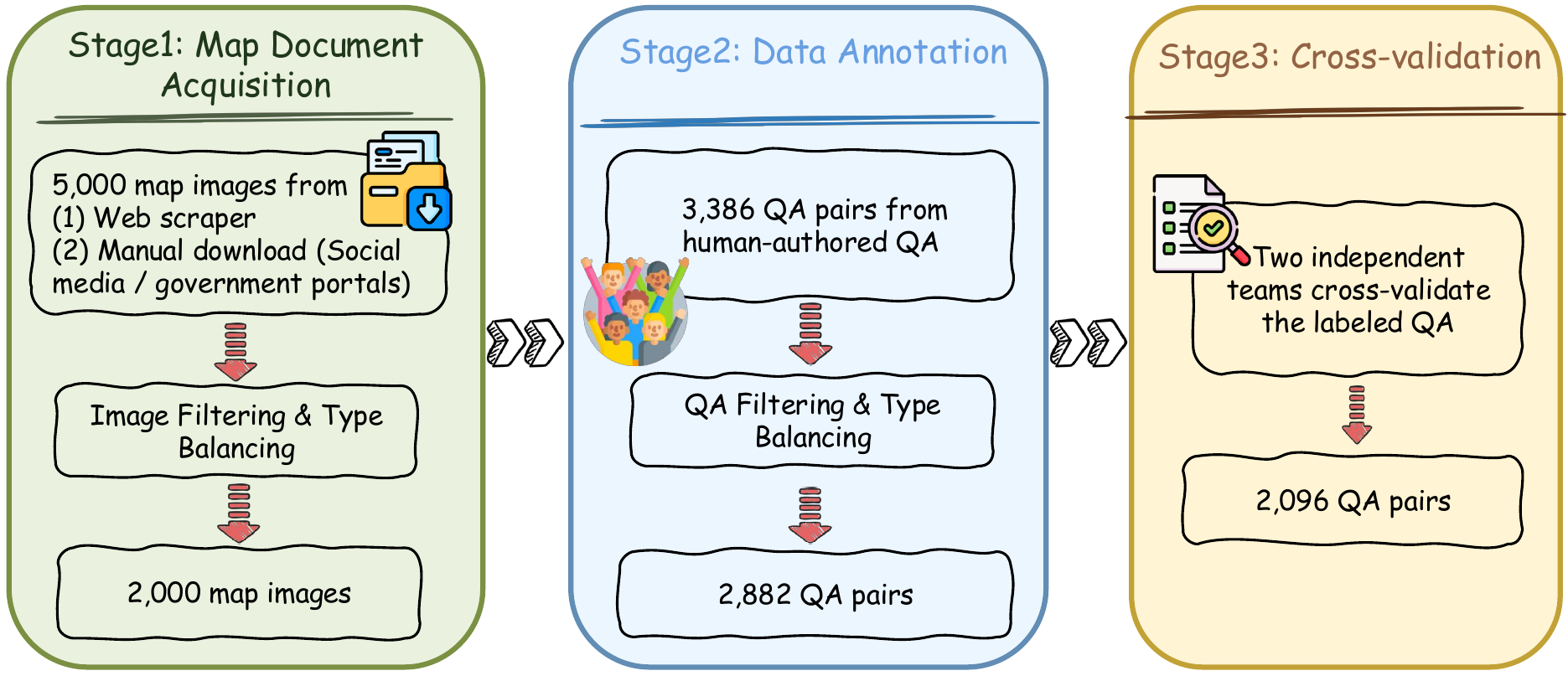}
\caption{Overview of the OmniMapBench dataset construction.}
\label{fig:construct}
\end{figure}

The construction of the OmniMapBench dataset is a meticulous, multi-stage process designed to ensure high diversity, quality, and reasoning complexity. As illustrated in Figure~\ref{fig:construct}, the pipeline is structured into three main stages: (1) Data Acquisition, where a broad collection of map images is gathered and filtered; (2) Data Annotation, where model-aided human experts create question-answer pairs; and (3) Cross-Validation, where the annotated pairs undergo rigorous verification to produce the final benchmark.

\begin{table}[t]
\centering
\caption{Statistics of the OmniMapBench dataset.}
\label{tab:dataset_statistics}
\footnotesize
\setlength{\tabcolsep}{3pt}

\begin{subtable}[t]{0.48\textwidth}
    \centering
    \caption{Image category distribution.}
    \label{tab:image_category_stats}
    \begin{tabular}{@{}lrr@{}}
    \toprule
    \textbf{Category} & \textbf{\#} & \textbf{\%} \\
    \midrule
    Indoor Navigation              & 294 & 18.34 \\
    Education \& & 276 & 17.22 \\
    \quad History & & \\
    Engineering \& & 269 & 16.78 \\
    \quad Facilities & & \\
    Transportation                 & 251 & 15.66 \\
    Tourism                        & 211 & 13.16 \\
    Topography \& & 117 & 7.30  \\
    \quad Remote Sensing & & \\
    Environment \& & 83 & 5.18  \\
    \quad Nature & & \\
    Socioeconomic                  & 63 & 3.93  \\
    Game                           & 39  & 2.43  \\
    \midrule
    \textbf{Total} & \textbf{1,603} & \textbf{100} \\
    \bottomrule
    \end{tabular}
\end{subtable}
\hfill
\begin{subtable}[t]{0.48\textwidth}
    \centering
    \caption{QA pair attributes.}
    \label{tab:qa_attribute_stats}
    \begin{tabular}{@{}llrr@{}}
    \toprule
    \textbf{Attribute} & \textbf{Category} & \textbf{\#} & \textbf{\%} \\
    \midrule
    \multirow{3}{*}{QA Type} 
        & Single Choice & 1,461 & 69.70 \\
        & Multi Choice  & 269  & 12.83 \\
        & Sort          & 366  & 17.46 \\
    \midrule
    \multirow{3}{*}{Capability} 
        & L1 (Perc.) & 441 & 21.04 \\
        & L2 (Single.)  & 802  & 38.26 \\
        & L3 (Multi.) & 853  & 40.70 \\
    \midrule
    \multirow{2}{*}{Language} 
        & English & 1,274 & 60.78 \\
        & Chinese & 822  & 39.22 \\
    \midrule
    \multicolumn{2}{@{}l}{\textbf{Total}} & \textbf{2,096} & \textbf{100} \\
    \bottomrule
    \end{tabular}
\end{subtable}

\end{table}

\subsection{Data Acquisition}
The primary goal of the acquisition stage is to assemble a large and visually diverse corpus of map documents. The process commences with the collection of over 5,000 map images. A dual-modality approach is employed to balance scale and quality. 
First, an automated web scraping procedure is executed. To ensure comprehensive coverage across the nine defined map categories, a LLM (GPT-5\cite{gpt5card}) is utilized to generate a wide array of search query variations from a set of seed keywords. 
Second, a targeted manual collection process is conducted to source high-quality, specialized maps. This is particularly important for categories such as transit maps and tourist diagrams, which demand high resolution and specific visual styles. These images are manually sourced from specialized platforms, including government portals and social media channels.

The initial corpus of 5,000 images is then subjected to a rigorous filtering protocol. Images are systematically removed based on three criteria: (1) high visual similarity to other images, indicating redundancy; (2) insufficient resolution or clarity, which would impede detailed analysis; and (3) content related to politically sensitive regions. After this screening, a refined set of approximately 2,000 high-quality map images is retained for the annotation stage.

\begin{table*}[t]
\centering
\caption{Visual Dependency Index (VDI) and Descriptive Saturation Index (DSI) across benchmarks and token budgets (evaluated with gpt-4.1-2025-04-14\cite{gpt4-1}). Values are rounded to three decimals. At $k_{\text{max}}=1024$, for VDI, the \textbf{maximum} value is bolded and the second maximum is underlined; for DSI, the \textbf{maximum} value is bolded and the minimum is underlined, highlighting benchmarks with distinct characteristics in visual dependency and descriptive complexity.}
\label{tab:vdi_dsi_combined}
\small
\begin{tabular}{l|cc|cc|cc|cc|cc}
\toprule
\textbf{Benchmark} & \multicolumn{2}{c|}{$k_{\max}=64$} & \multicolumn{2}{c|}{$k_{\max}=128$} & \multicolumn{2}{c|}{$k_{\max}=256$} & \multicolumn{2}{c|}{$k_{\max}=512$} & \multicolumn{2}{c}{$k_{\max}=1024$} \\
\cmidrule(lr){2-3} \cmidrule(lr){4-5} \cmidrule(lr){6-7} \cmidrule(lr){8-9} \cmidrule(lr){10-11}
& VDI & DSI & VDI & DSI & VDI & DSI & VDI & DSI & VDI & DSI \\
\midrule
AI2D~\cite{hiippala2021ai2d}           & 0.892 & 0.780 & 0.613 & 0.790 & 0.159 & 0.745 & 0.035 & 0.597 & 0.019 & 0.336 \\
ChartQA~\cite{masry2022chartqa}        & 0.559 & 0.842 & 0.343 & 0.891 & 0.183 & 0.847 & 0.147 & 0.698 & \underline{0.143} & 0.394 \\
DocVQA~\cite{mathew2021docvqa}         & 0.552 & 0.855 & 0.357 & 0.944 & 0.224 & 0.920 & 0.122 & 0.795 & 0.076 & 0.516 \\
InfoVQA~\cite{mathew2022infographicvqa}        & 0.517 & 0.897 & 0.352 & 0.990 & 0.241 & 0.985 & 0.155 & 0.961 & 0.049 & \textbf{0.762} \\
MathVista~\cite{lu2023mathvista}      & 0.882 & 0.722 & 0.682 & 0.722 & 0.347 & 0.617 & 0.126 & 0.478 & 0.051 & 0.269 \\
MMMU~\cite{yue2024mmmu}           & 0.917 & 0.676 & 0.841 & 0.727 & 0.526 & 0.669 & 0.221 & 0.503 & 0.093 & 0.293 \\
MMMU-Pro~\cite{yue2025mmmu}       & 0.904 & 0.666 & 0.848 & 0.710 & 0.578 & 0.615 & 0.257 & 0.459 & 0.117 & 0.273 \\
MMBench~\cite{liu2024mmbench}        & 0.822 & 0.591 & 0.510 & 0.629 & 0.114 & 0.518 & 0.035 & 0.364 & 0.028 & 0.195 \\
OCRBench~\cite{liu2024ocrbench}       & 0.222 & 0.507 & 0.149 & 0.492 & 0.095 & 0.378 & 0.044 & 0.244 & 0.043 & 0.124 \\
OCRBench-v2~\cite{fu2024ocrbench}    & 0.411 & 0.781 & 0.282 & 0.861 & 0.160 & 0.813 & 0.108 & 0.680 & 0.085 & 0.437 \\
\midrule
\textbf{OmniMapBench}   & 0.556 & 0.911 & 0.527 & 0.983 & 0.520 & 0.963 & 0.474 & 0.915 & \textbf{0.338} & \underline{0.629} \\
\bottomrule
\end{tabular}
\end{table*}

\subsection{Annotation and Validation}
The annotation task is performed by professional human annotators. Human annotators are explicitly instructed to create \textbf{map-related challenging QA pairs}. To further standardize the evaluation and minimize ambiguity, QA pairs are formulated in objective formats, like single-choice, multiple-choice, or ranking questions. Strict guidelines are established for question creation. Each question must be answerable using only the visual information present in the map, without reliance on external knowledge. Annotators are also required to craft questions that span three predefined capability levels: basic perception (Level 1), single-step spatial reasoning (Level 2), and complex, multi-step relational reasoning (Level 3). This stage initially yields approximately 3,386 QA pairs. This collection then undergoes a preliminary filtering and type-balancing step, which reduces the set to 2,882 pairs for validation.

 To ensure maximum quality and correctness, the 2,882 candidate QA pairs are subjected to a rigorous cross-validation process. The validation is carried out by two independent teams of annotators who did not participate in the initial annotation of the data they review. Each team assesses every QA pair for clarity, correctness of the answer, and adherence to the specified capability level. Any pair that is deemed ambiguous, incorrect, or miscategorized by either team is discarded. This multi-layered verification protocol effectively eliminates potential errors and subjective biases. The entire construction pipeline concludes with a final set of approximately 2,096 meticulously verified QA pairs, which form the OmniMapBench dataset.

\subsection{Dataset Statistics}
Table~\ref{tab:dataset_statistics} presents a detailed breakdown of the OmniMapBench statistics. The image collection is composed of 1,603 images distributed across nine categories. The largest concentrations of images are found in Indoor Navigation (18.34\%), Education \& History (17.22\%), and Engineering \& Facilities (16.78\%). The remaining categories contribute to a wide-ranging visual diversity, from schematic diagrams to photorealistic scenes, ensuring comprehensive coverage. Furthermore, all images are retained at their native resolutions to preserve the authentic diversity and fine-grained visual density of real-world maps. As a result, the image sizes vary broadly from $233\times 464$ to $11{,}811\times 9{,}442$ pixels, with a mean resolution of $2{,}055\times 1{,}705$. The question-answering component contains 2,096 pairs. These are distributed by type into single-choice (69.70\%), multiple-choice (12.83\%), and sorting (17.46\%). In terms of capability levels, questions are classified as Level 3 (Multi-step reasoning) at 40.70\%, Level 2 (Single-step reasoning) at 38.26\%, and Level 1 (Perception) at 21.04\%. The language distribution consists of 60.78\% English and 39.22\% Chinese, which facilitates cross-lingual evaluation.

\subsection{Comparison with other Benchmarks}

A high-level contrast is provided in Table~\ref{tab:compare}. Prior document-understanding benchmarks primarily focus on content such as industrial documents, charts, or OCR-rich images that are often reducible to text. More recent map-specific benchmarks have emerged, but they tend to specialize in a single map type, such as transit maps (\eg, ReasonMap\cite{feng2026reasonmap}) or choropleth maps (\eg, MapQA\cite{chang2022mapqadatasetquestionanswering}). In contrast, OmniMapBench encompasses 9  map categories with diverse visual styles and manually annotated QA pairs. Comparison of visual dependency is shown in Sec.~\ref{sec:vdi}.

\begin{table}[h]
\centering
\caption{High-level comparison with other benchmarks. }
\label{tab:compare}
\begin{tabular}{lccc}
\hline
Benchmark & Document Type & Cases & Images \\
\hline
DocVQA\cite{mathew2021docvqa}      & Industrial document & \textasciitilde50k  & \textasciitilde12k \\
ChartQA\cite{masry2022chartqa}     & Chart                & \textasciitilde2.5k  & \textasciitilde1.6k  \\
AI2D\cite{hiippala2021ai2d}        & Science diagram      & \textasciitilde15k  & \textasciitilde5k \\
OCRBenchV2\cite{fu2024ocrbench}  & OCR-rich image    & \textasciitilde10k & \textasciitilde1.5k \\
InfoVQA\cite{mathew2022infographicvqa}     & Infographic          & \textasciitilde3k & \textasciitilde0.5k \\
ReasonMap\cite{feng2026reasonmap}     & Transit map          & \textasciitilde1k & \textasciitilde30\\
MapQA\cite{chang2022mapqadatasetquestionanswering}     & Choropleth map          & \textasciitilde800k & \textasciitilde60k\\
MapQA\cite{li2025mapqa}     & OpenStreetMap          & \textasciitilde3k & / \\
\hline
OmniMapBench & Diverse maps                 & 2,096  & 1,603 \\
\hline
\end{tabular}
\end{table}

\section{Visual Dependency Evaluation}
\label{sec:vdi}

\begin{figure*}[t]
\centering
\includegraphics[width=0.99\linewidth]{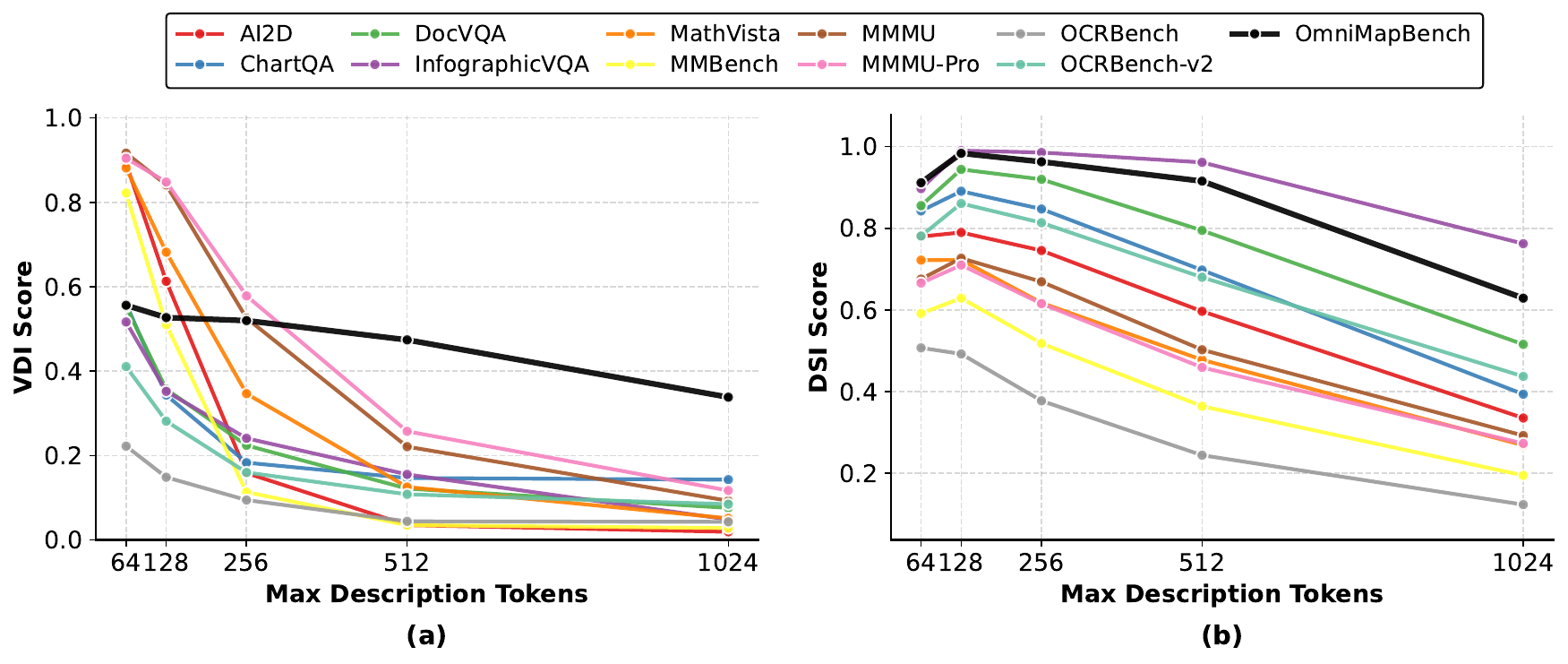}
\caption{Visualization of VDI (a) and DSI (b) across increasing description token budgets.}
\label{fig:vdi}
\end{figure*}

A large portion of existing VQA benchmarks is solvable through text-mediated shortcuts. In this setting, the direct pathway \texttt{(Image, Question) -> Answer} is replaced by first producing a textual question-agnostic description of the image and then answering from text. High accuracy is often achieved with short descriptions, which indicates predominantly linguistic reasoning and limited visual dependency. To quantify this reliance on vision, the Visual Dependency Index (VDI) is proposed as the normalized accuracy drop when image input is replaced by a textual surrogate under a token budget. A higher VDI indicates stronger dependence on visual cues that are not easily captured by text. Formal definitions are provided next.

\subsection{Formal Definitions}
\label{ssec:vdi_definition}

To ensure a fair and standardized comparison across different benchmarks, the evaluation model is held constant. A single, powerful LVLM, denoted as $\mathcal{M}$, is selected to serve as the reference for measuring visual dependency. Let a benchmark be $\mathcal{B} = \{(I_i, Q_i, A_i)\}_{i=1}^N$, where $I_i$ is an image, $Q_i$ is a question, and $A_i$ is the ground-truth answer. The standard performance of the reference model $\mathcal{M}$ on benchmark $\mathcal{B}$ is its accuracy on the direct inference task. This baseline, denoted as $\text{Acc}_{\text{full}}(\mathcal{B})$, is defined as:
\begin{equation}
\label{eq:acc_full}
\text{Acc}_{\text{full}}(\mathcal{B}) = \frac{1}{N} \sum_{i=1}^N \mathbb{I}(\mathcal{M}(I_i, Q_i) = A_i)
\end{equation}

For the text-mediated pathway, the same model $\mathcal{M}$ is used to generate a description $D_{i, k_{\text{max}}}$ for each image $I_i$, constrained by a maximum token budget of $k_{\text{max}}$. Subsequently, $\mathcal{M}$ is queried again, this time operating in a language-only mode with an input composed of the generated description $D_{i, k_{\text{max}}}$ and the original question $Q_i$. The text-only accuracy on benchmark $\mathcal{B}$, denoted as $\text{Acc}_{\text{text}}(\mathcal{B}, k_{\text{max}})$, is then:
\begin{equation}
\label{eq:acc_text}
\text{Acc}_{\text{text}}(\mathcal{B}, k_{\text{max}}) = \frac{1}{N} \sum_{i=1}^N \mathbb{I}(\mathcal{M}(D_{i, k_{\text{max}}}, Q_i) = A_i)
\end{equation}

The \textbf{Visual Dependency Index (VDI)} for a given benchmark $\mathcal{B}$, measured with respect to the reference model $\mathcal{M}$ and a token budget $k_{\text{max}}$, is defined as the normalized performance drop:
\begin{equation}
\label{eq:vdi}
\text{VDI}(\mathcal{B}, k_{\text{max}}) = \frac{\text{Acc}_{\text{full}}(\mathcal{B}) - \text{Acc}_{\text{text}}(\mathcal{B}, k_{\text{max}})}{\text{Acc}_{\text{full}}(\mathcal{B})}
\end{equation}

By fixing $\mathcal{M}$, VDI$(\mathcal{B}, k_{\text{max}})$ becomes a direct property of the benchmark $\mathcal{B}$ itself, quantifying its inherent dependency on visual information that resists textual summarization under a specific resource constraint $k_{\text{max}}$. The VDI score ranges from approximately 0 to 1. A score near 0 indicates that the task can be almost entirely solved using textual information, while a score near 1 indicates a critical dependency on visual input.

A key methodological consideration in this process is that generative models operate with a maximum token limit ($k_{\text{max}}$), not a fixed output length. The actual number of generated tokens, $k_{\text{actual}}$, is a variable that depends on both the model and the complexity of the image content. To analyze this behavior and to provide a complementary view on benchmark complexity, the \textbf{Descriptive Saturation Index (DSI)} is introduced as a secondary metric. DSI quantifies the model's tendency to utilize its allocated token budget.

The DSI for a benchmark $\mathcal{B}$ at a given $k_{\text{max}}$ is defined as the average ratio of the actual tokens generated to the maximum allowed:
\begin{equation}
\label{eq:dsi}
\text{DSI}(\mathcal{B}, k_{\text{max}}) = \mathbb{E}_{i \sim \mathcal{B}} \left[ \frac{k_{\text{actual}, i}}{k_{\text{max}}} \right]
\end{equation}
where $k_{\text{actual}, i}$ is the length of the description generated for the $i$-th sample. A DSI value close to 1 suggests the benchmark's content is descriptively dense, consistently requiring the full token budget. A lower DSI indicates the content is more easily summarizable. Together, VDI and DSI provide a more comprehensive characterization of a benchmark's demands on visual reasoning.

\subsection{Experimental Setup}

\paragraph{Benchmarks.}

The evaluation protocol is applied to OmniMapBench and ten established visual reasoning benchmarks, as detailed in Table~\ref{tab:vdi_dsi_combined}. To ensure a comprehensive comparison, this selection spans multiple domains. It includes benchmarks focused on document understanding (\textit{e.g.}, DocVQA\cite{mathew2021docvqa}), chart reasoning (\textit{e.g.}, ChartQA\cite{masry2022chartqa}), mathematical problem-solving (\textit{e.g.}, MathVista), and broad multi-disciplinary reasoning (\textit{e.g.}, MMMU\cite{yue2025mmmu}, AI2D\cite{hiippala2021ai2d}).

\paragraph{Procedure and Models.}
The procedure for deriving VDI and DSI curves is outlined in Algorithm~\ref{alg:vdi_dsi_computation_generalized}. A benchmark $\mathcal{B}$ with $N$ samples is considered. A single reference LVLM, \texttt{gpt-4.1-2025-04-14\cite{gpt4-1}}, is fixed as $\mathcal{M}$ for all stages: (i) obtaining $\text{Perf}_{\text{full}}$ via direct visual inference on $(I_i, Q_i)$; (ii) generating descriptions $D_{i,k_{\max}}$ under token budgets $K=\{64,128,256,512,1024\}$; and (iii) performing text-only inference with $(D_{i,k_{\max}}, Q_i)$. For each $k_{\max}$, the actual token length $k_{\text{actual},i}$ is recorded and exact-match accuracy $\text{Acc}_{\text{text}}(k_{\max})$ is computed. VDI and DSI are then computed following Equations~\ref{eq:vdi} and~\ref{eq:dsi}. The prompts for generating descriptions without $Q_i$ and answering with $Q_i$ are shown in Figure~\ref{fig:prompt_description} and Figure~\ref{fig:prompt_text_only}.

\begin{figure*}[!ht]
\begin{AIbox}{Prompt for Generating Image Descriptions}
{
Please provide a text description for this image that is as detailed as possible. You can also use the corresponding code to describe information such as charts/tables. No more than \{{max\_tokens\}} tokens. Only output the description, no other text.
}
\end{AIbox}
\caption{The prompt used to generate question-agnostic image descriptions for the VDI analysis}
\label{fig:prompt_description}
\end{figure*}

\begin{figure*}[!ht]
\begin{AIbox}{Prompt for Answering from Description}
{
\{{Question\}}.
Now, Please answer the question only based on the following image description: \{{image\_description\}}
}
\end{AIbox}
\caption{The prompt used for the text-only answering stage of the VDI analysis. It forces the model to rely exclusively on the generated text description, thereby measuring performance without direct visual input.}
\label{fig:prompt_text_only}
\end{figure*}

\begin{algorithm}[h]
\caption{Pseudocode for the Visual Dependency Evaluation Experiment Procedure}
\label{alg:vdi_dsi_computation_generalized}
\begin{algorithmic}[1]
\State \textbf{Input:} Benchmark $\mathcal{B} = \{(I_i, Q_i, A_i)\}_{i=1}^N$, VLM $\mathcal{M}$,
\Statex Baseline performance $\text{Perf}_{\text{full}}$, Token budgets $K = \{k_{\text{max}}^{(1)}, \dots, k_{\text{max}}^{(m)}\}$
\Statex
\For{each $k_{\text{max}} \in K$}
    \State Initialize $S_{\text{text}} \gets 0$ and $L_{\text{total}} \gets 0$
    \For{$i = 1$ to $N$}
        \State $D_{i, k_{\text{max}}} \gets \mathcal{M}.\text{GenerateDescription}(I_i, k_{\text{max}})$
        \State $k_{\text{actual}, i} \gets \text{Length}(D_{i, k_{\text{max}}})$
        \State $L_{\text{total}} \gets L_{\text{total}} + k_{\text{actual}, i}$
        \State $\hat{A}_{i} \gets \mathcal{M}(D_{i, k_{\text{max}}}, Q_i)$
        \State $S_{\text{text}} \gets S_{\text{text}} + \mathbb{I}(\hat{A}_{i} = A_i)$
    \EndFor
    \State $\text{Acc}_{\text{text}}(k_{\text{max}}) \gets S_{\text{text}} / N$
    \State $\text{VDI}(k_{\text{max}}) \gets (\text{Perf}_{\text{full}} - \text{Acc}_{\text{text}}(k_{\text{max}})) / \text{Perf}_{\text{full}}$
    \State $\text{DSI}(k_{\text{max}}) \gets (L_{\text{total}} / N) / k_{\text{max}}$
\EndFor
\Statex
\State \textbf{Output:} VDI and DSI curves parameterized by $K$
\end{algorithmic}
\end{algorithm}

\subsection{Evaluation Results and Analysis}
The empirical results for the VDI and DSI are presented in Figure~\ref{fig:vdi} and Table~\ref{tab:vdi_dsi_combined}. VDI decreases with larger token budgets across benchmarks. A slower decay is observed for OmniMapBench; at $k_{\text{max}}=1024$, a VDI of 0.338 is registered, the highest among all benchmarks. In contrast, benchmarks such as AI2D and MMBench drop to near-zero VDI (0.019 and 0.028), indicating that long descriptions largely eliminate visual dependency for them.

A complementary view is provided by DSI in Figure~\ref{fig:vdi}(b). OmniMapBench maintains high saturation across budgets, remaining at 0.629 at $k_{\text{max}}=1024$, the second highest at the maximum budget. This behavior indicates information-dense visual content that compels extensive token utilization, whereas several benchmarks (e.g., OCRBench at 0.124 and MMBench at 0.195) exhibit much lower saturation at large budgets. Taken together, the high DSI and VDI validate that OmniMapBench resists textual reduction and requires genuine visual-centric reasoning.

\subsection{Ablation on the Reference Model}

To assess sensitivity to the choice of the reference model, an ablation is conducted by replacing $M$ with another LVLM. The results demonstrate consistent trends across different reference models. 

\section{Benchmarks}

\begin{table*}[t]
\centering
\caption{Main evaluation results on OmniMapBench. Models are grouped by category and sorted by Final Score. The best result is \textbf{bolded}, and the second-best is \underline{underlined}.}
\label{tab:main_results_updated}
\begin{tabular}{llcccc}
\toprule
\textbf{Category} & \textbf{Model} & \textbf{Level 1} & \textbf{Level 2} & \textbf{Level 3} & \textbf{Final Score} \\
\midrule
Proprietary & Gemini-3.1-Pro~\cite{gemini31pro} & 88.24 & 81.53 & 62.88 & \textbf{75.03} \\
MLLMs & Gemini-2.5-Pro~\cite{gemini25pro} & 73.70 & 70.70 & 46.31 & 61.40 \\
 & Doubao-seed-1.6-vision~\cite{seed16} & 72.94 & 63.65 & 41.80 & 56.65 \\
 & GPT-5~\cite{gpt5card} & 63.10 & 61.47 & 48.53 & 56.54 \\
 & GPT-5-mini~\cite{gpt5card} & 65.60 & 62.59 & 45.13 & 56.11 \\
 & GPT-4.1-mini~\cite{gpt4-1} & 60.59 & 47.13 & 35.64 & 45.27 \\
 & Claude-Sonnet-4-5~\cite{claudesonnet4-5} & 58.93 & 52.63 & 31.29 & 45.04 \\
 & GPT-4.1~\cite{gpt4-1} & 48.52 & 45.14 & 37.87 & 42.88 \\
 & Claude-Opus-4-1~\cite{claudeopus4-1} & 56.12 & 44.47 & 33.50 & 42.25 \\
 & GPT-5-nano~\cite{gpt5card} & 45.56 & 40.52 & 32.83 & 38.44 \\
 & GPT-4o~\cite{gpt4o} & 40.09 & 31.80 & 26.73 & 31.47 \\
\midrule
Open-Source & Qwen3.5-397B-A17B~\cite{qwen35} & 83.89 & 79.01 & 60.37 & \underline{72.22} \\
MLLMs & Qwen3.5-Plus~\cite{qwen35} & 85.32 & 77.64 & 57.47 & 70.80 \\
 & Qwen3.5-27B~\cite{qwen35} & 82.93 & 77.10 & 53.97 & 68.62 \\
 & Qwen3.5-122B-A10B~\cite{qwen35} & 83.13 & 76.52 & 53.54 & 68.37 \\
 & Kimi-K2.5~\cite{kimiteam2026kimik25visualagentic} & 74.77 & 74.80 & 56.71 & 67.64 \\
 & Qwen3.5-35B-A3B~\cite{qwen35} & 81.10 & 72.27 & 49.93 & 64.75 \\
 & Qwen3.5-Flash~\cite{qwen35} & 79.27 & 73.87 & 48.62 & 64.56 \\
 & Gemma-4-31B-it~\cite{gemma4} & 60.77 & 58.35 & 40.68 & 51.67 \\
 & GLM-4.5V~\cite{vteam2025glm45vglm41vthinkingversatilemultimodal} & 66.67 & 54.24 & 35.80 & 49.33 \\
 & InternVL3.5-241B-A28B~\cite{wang2025internvl35advancingopensourcemultimodal} & 68.48 & 52.99 & 30.36 & 47.04 \\
 & Intern-S1~\cite{bai2025intern} & 61.22 & 49.13 & 32.71 & 44.99 \\
 & Gemma-4-26B-A4B-it~\cite{gemma4} & 51.25 & 49.75 & 32.83 & 43.18 \\
 & Qwen3-VL-8B~\cite{qwen3vl} & 60.09 & 44.89 & 29.07 & 41.65 \\
 & Gemma-3-27B-it~\cite{gemma3} & 34.92 & 31.80 & 25.44 & 29.87 \\
\bottomrule
\end{tabular}%
\end{table*}

This section details the comprehensive evaluation of leading LVLMs on the OmniMapBench dataset. 

\subsection{Setup}
\paragraph{Evaluated Models.}
A comprehensive suite of 25 representative LVLMs is evaluated to provide a broad view of the current landscape. These models are categorized into proprietary and open-source systems. The proprietary models include leading LVLMs such as GPT-5, Gemini-3.1-Pro. The open-source selection encompasses prominent models like Qwen3.5 and GLM series. A complete list of all evaluated models is provided in Table \ref{tab:main_results_updated}.

\paragraph{Evaluation Protocol.}
For all question types (single-choice, multiple-choice, and sorting), a strict accuracy metric is employed. A model's response is considered correct only if it exactly matches the ground-truth. To elicit the maximum reasoning capabilities of the models, a Chain-of-Thought\cite{wei2023chainofthoughtpromptingelicitsreasoning} prompting strategy is applied.

\subsection{Results and Analysis}
\begin{figure}
    \centering
    \includegraphics[width=0.75\linewidth]{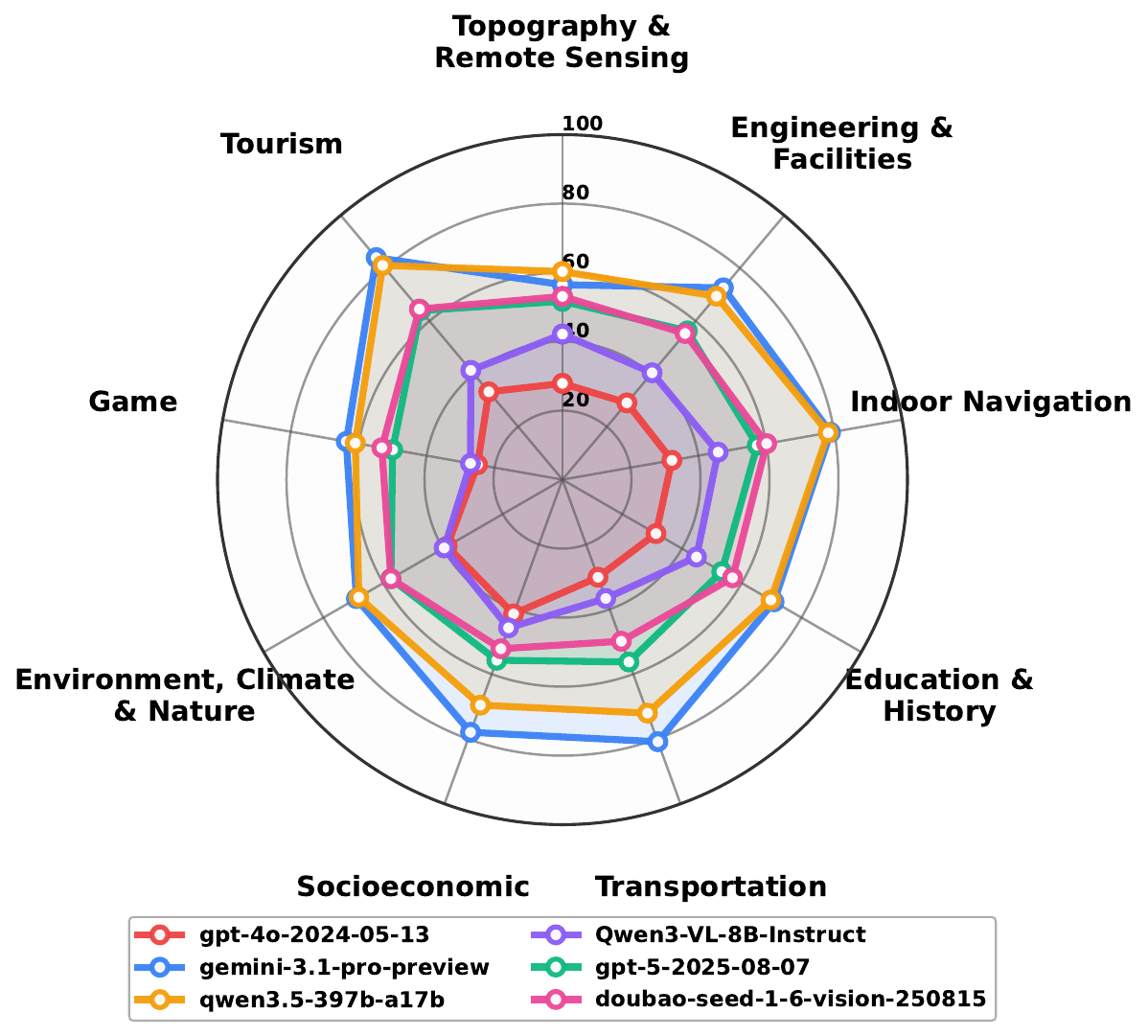}
    \caption{Radar chart comparing the performance of six representative models across nine map categories in OmniMapBench.}
    \label{fig:radar}
\end{figure}

The primary evaluation results are presented in Table \ref{tab:main_results_updated} and Figure \ref{fig:radar}. Table \ref{tab:main_results_updated} decomposes performance across three capability levels, while Figure \ref{fig:radar} provides a category-specific breakdown across the nine map types.

\paragraph{Overall Performance.}
A pronounced disparity across models is observed. The best final accuracy is achieved by Gemini-3.1-Pro at 75.03\%. Among open-source models, Qwen3.5-397B-A17B attains the highest accuracy at 72.22\%, demonstrating high competitiveness with the top proprietary system, with a narrow gap of less than 3 percentage points.  Notably, the leading open-source models, such as the Qwen3.5 series, significantly outperform the majority of other proprietary models listed, challenging the notion that proprietary systems consistently hold a significant advantage. This indicates that while OmniMapBench remains a difficult benchmark, the gap in visual-centric reasoning between the best open-source and proprietary models is closing.

\paragraph{No-Image Baseline.}
To assess whether OmniMapBench can be solved by relying on language priors, world knowledge, or answer-choice bias, a no-image ablation is conducted. In this setting, only the textual question and the corresponding answer choices are provided to the models, without any visual input or textual image description. Three representative models are evaluated under this protocol. A substantial accuracy drop is observed, with the average performance decreasing from 58.87\% in the standard setting to 23.35\% in the blind setting. This degradation indicates that OmniMapBench cannot be reliably solved through textual shortcuts or inherent biases alone.

\paragraph{Performance by Capability Level.}

A clear trend of declining performance with increasing reasoning complexity is observed, from Level 1 (Perception) to Level 3 (Multi-step reasoning). For instance, Gemini-3.1-Pro records accuracies of 88.24\%, 81.53\%, and 62.88\% at Levels 1, 2, and 3, respectively. These results indicate that while models handle basic perception and direct information extraction relatively well, multi-step spatial and visual-centric reasoning remains the principal bottleneck.

\paragraph{Performance by Map Category.}
Figure \ref{fig:radar} reveals considerable performance variance across different map types. This suggests that model capabilities are not uniform across the diverse visual domains represented in OmniMapBench. Performance is generally higher on map types that contain more structured or conventional textual information, such as \textit{Tourism}. In contrast, lower scores are observed for categories heavily reliant on the interpretation of dense, non-standard symbols and complex topological relationships, like \textit{Topography \& Remote Sensing}. This finding indicates a direction for future work in improving the LVLMs.

\section{Conclusion}
This work introduces OmniMapBench, a visual-centric benchmark for map document understanding, encompassing 2,096 manually verified QA pairs across 9 categories and spanning perception to multi-step reasoning tasks. The Visual Dependency Index (VDI) is proposed to quantify resistance to textual reduction under token budgets. Comprehensive evaluations reveal that state-of-the-art LVLMs achieve 75.03\% accuracy, with high residual VDI confirming the benchmark's reliance on genuine visual grounding. The findings expose significant performance gaps in visual-centric reasoning. The dataset, metrics, and protocol provide a principled framework for assessing visual-centric reasoning of map documents. It is hoped that this benchmark catalyzes progress toward more robust multimodal reasoning and scalable perception-reasoning integration.



%
%
\bibliographystyle{splncs04}
\bibliography{main}

\end{document}